# A Novel Prediction Approach for Exploring PM2.5 Spatiotemporal Propagation Based on Convolutional Recursive Neural Networks

Hsing-Chung Chen[1,2]

[1] *Department of Computer Science and Information Engineering, Asia University, Taiwan*
[2] *Department of Medical Research, China Medical University Hospital, China Medical University, Taiwan*

Karisma Trinanda Putra[1,3]

[3] *Department of Electrical Engineering, Universitas Muhammadiyah Yogyakarta, Indonesia*

Jerry Chun-WeiLin[4]

[4] *Department of Computer Science, Electrical Engineering and Mathematical Sciences, Western Norway University of Applied Sciences, Norway*

---

**Abstract**

The spread of PM2.5 pollutants that endanger health is difficult to predict because it involves many atmospheric variables. These micron particles can spread rapidly from their source to residential areas, increasing the risk of respiratory disease if exposed for long periods. The prediction system of PM2.5 propagation provides more detailed and accurate information as an early warning system to reduce health impacts on the community. According to the idea of transformative computing, the approach we propose in this paper allows computation on the dataset obtained from massive-scale PM2.5 sensor nodes via wireless sensor network. In the scheme, the deep learning model is implemented on the server nodes to extract spatiotemporal features on these datasets. This research was conducted by using dataset of air quality monitoring systems in Taiwan. This study presents a new model based on the convolutional recursive neural network to generate the prediction map. In general, the model is able to provide accurate predictive results by considering the bonds among measurement nodes in both spatially and temporally. Therefore, the particulate pollutant propagation of PM2.5 could be precisely monitored by using the model we propose in this paper.

---

## 1. Introduction

Particulate matter (PM) below 2.5 um, which is called PM2.5, is one of the largest and most dangerous sources of air pollutants., PM2.5 is very dangerous for health especially for vulnerable groups, such as infants, children, pregnant women, and the elderly, etc. High PM2.5 concentrations often cause to lots of diseases such as heart disease, respiratory infections, cancer, and chronic lung disease [1]. PM2.5 is a micro particle that could aggravate the respiratory disease more quickly because it could settle to the respiratory tract of the bronchi and alveoli [2]. PM2.5 is more dangerous than particulate matters below 10 (PM10) because it is attached to the lung bubbles instead of filtering in the upper respiratory system. Thus, it could reduce the ability of the lungs in gas exchange. Due to its very light physical characteristics and odorless, the long-term various unhealthy impacts will be less and less perceived by the residents [3].

There are many factors that greatly affect the measurement of this pollutant sources. PM2.5 pollution data has the characteristics which are difficult to precisely detecting, high data randomness, and susceptible changing compared with other atmospheric variables [4]. PM2.5 sensors require a certain period time for measurements in order to generate stable as well as precise results. Meanwhile, the data measured by the PM2.5 sensors has a randomness level owing to the various height and location of the sensors. Moreover, the ever-changing PM2.5 variable is difficult to predict because the features are highly influenced by other variables *e.g.* geographical location, wind direction, temperature, humidity, and other sources of pollutants [5]. Measuring air pollution on a massive scale deployed sensors is even more difficult challenges. The use of a large number of sensors actually increases the chance of data randomness. The large amount of data generated from thousands of sensors creates confusion for users if the data is directly presented in real-time. There is even a possibility that some measurement errors are obtained from millions of data, resulting in inconsistencies in the results. However, the spatial pattern of air pollutant distribution could be monitored in more detail. The collected data also has a sequential pattern feature that changes over time. These features of the collected data could be learned by an Artificial Intelligence (AI) model so that they could be used to build an accurate prediction model. The prediction model based on the spatiotemporal pattern will be theoretically improved the accuracy because, each data tied together in terms of space and time domain [6-8].

We use a large number of data captured from massive-scale wireless sensor networks (WSN) in Taiwan to provide earlier and more accurate predictions. One of the weaknesses of massive-scale WSN is the large amount of data that is collected and neglected. Sometimes the data is deleted to reduce the storage burden. Transformative computing (TC) [9] has become a novel issue that could be used to solve this problem mentioned above. TC is a new paradigm that has the following advantages. First, it is a fast computing approach without wasting data, which increase the network load. Second, TC provides the possibility to combine several existing information technologies. Finally, TC is expected to increase the efficiency of data exploration as well as analysis, and decision making. All of these targets could be achieved by implementing an AI model *e.g.* deep learning into WSN. On the other hand, Taiwan as a developed country especially in the information technology area has funded many projects related to WSN application. The Taiwanese government has implemented PM-25 measuring sensors in every weather station distributed in each district. Cooperation between the government and WSN manufacturers has also generated many sensor nodes installed with a large but uneven distribution. This collaboration generates a pollutant measurement system *i.e.* 'Airbox' devices [10] that have been implemented and could be accessed by the public. The 'Airbox' is a PM2.5 measurement project with community support. This device is designed according to industry standards, but its implementation is carried out by the wider community. These sensor nodes take advantage of distributed WSN technology with a centralized monitoring system on the server. The data is recorded within a certain period and it could be accessed in real-time via the internet.

The spatiotemporal pattern could be used to predict the distribution of pollutants in the future [11]. This prediction is useful as an early warning system, because of its huge impact on public health. Furthermore, the direction of pollutant propagation could be predicted so that it could be used to build an accurate early warning system and could serve as a guide for the government to determine public policies. We use a Recursive Neural Network (RNN) modeling to record short and long-term patterns from the PM2.5 dataset obtained from the 'Airbox' devices. In addition, the spatial patterns on the dataset learned by using Convolutional Neural Network (CNN) modeling. The combination of CNN and RNN in a deep learning model has the ability to record many and complex pre-trained patterns [12]. This training process generates knowledge on the AI model which could be used to predict the propagation of PM2.5 from time to time. Moreover, TC could be used to build a base schema that utilized AI on large-scale sensor networks [13]. On the schema, the dataset and prediction results could be accessed by the public. It means that large WSNs with many variations of sensors work together in a system that could be accessed by researchers, students, or even the community to generate prediction models, policies, and public awareness that could improve the quality of public health.

The contributions of this paper are summarized below.

1) The novel predictor model capable of servicing large-scale sensor networks with thousands of sensor nodes is developed to generate high-resolution PM2.5 propagation maps.
2) This framework utilizes CNN to extract spatial features of a WSN and RNN to learn the temporal features from data sequences so that the correlation relationship between measurement results could be maintained. Moreover, the spatiotemporal correlation relationship is extracted as feature in order to provide better forecasting results than the previous approach.
3) This framework supports reconfigurable network deployment, *e.g.* reduction, addition, or topology modification to increase measurement precision without the need for repetitive training processes.
4) The quantitative analyses in terms of prediction accuracy and error rate are performed by using the real-time 'Airbox' dataset obtained from 268 sensor nodes located in Central Taiwan. We also present a qualitative analysis of the predicted results.
5) The geospatial heat maps that describe predictions of the air pollutants propagation are provided for the next day, where the prediction model only uses PM2.5 data variables without involving other atmospheric variables.

The remainder of this paper is described as follows. Section II reviews the previous works in terms of PM2.5 prediction. Section III provides basic methods to build the prediction model. Section IV describes in detail our proposed model. Section V shows the experiment and the result. Furthermore, we provide a discussion in Section VI. Finally, we summarize our conclusions in Section VII.

## 2. Related Works

PM2.5 is a micro particle that floats easily in Earth's atmosphere. The propagation of PM2.5 is closely related to changes in atmospheric variables, *e.g.* wind speed and direction. Wind speed and direction data have a high degree of randomness and always change over different periods [14]. It affects the direction, area, and speed of this pollutant propagation [15]. Until now, there are only a few models that describe the propagation of micro particles in a predictor model that includes time and space domains [16-18]. Furthermore, the existing models have not been applied to massive scale sensor networks. The probability of the error rate increases, with the increasing number of sensor nodes installed on a system [19]. Especially if the measurement area is very large with an uneven distribution of sensor nodes. Another challenge is that this measurement always changes depending on temperature, humidity, and geographical position. To overcome this problem, several studies that predict PM2.5 concentration use clustered data to simplify the process because it involves massive data from thousands of sensor nodes. By using clustering methods, the data processed is much less. It is very useful for large-scale predictor modeling, at the expense of the resolution of the prediction results. Linear regression analysis could be used as a method for making forecasting models with not too many clusters. However, this method cannot capture too many and complex features [20]. The longer the time to be predicted, the more different the prediction results from the ground truth.

Several researchers [21, 22] utilized Neural Networks (NN) to model signal complexity with a high degree of randomness.

However, the results still use a small 1-dimensional dataset without spatiotemporal feature extraction. Other researchers have even used Extreme Machine Learning (ELM) to predict air pollution [23]. However, this approach could only be used for small dataset with features which are not too complex. The more complex the learned features, Neural Network is unable to achieve convergent learning. Moreover, these features are stored in data sequences for a certain period. The discovery of the Recursive Neural Network model raises new possibilities related to forecasting sequential data even for high complexity features [24]. Moreover, entering the era of deep learning, many predictive models are starting to emerge and it produces better accuracy than conventional NN.

Multiple variants of RNN *e.g.* Gated Recurrent Unit (GRU) network and Long Short-Term Memory (LSTM) network are proven to be used to build sequential data prediction systems [25, 26]. The temporal pattern shows a very strong data correlation between the present data and the previous data. This correlation learned by GRU and LSTM networks and becomes knowledge *i.e.* manifested in the convergent weights on each layer. In addition, data measurement on a WSN has the features in the form of spatial and temporal correlation. The temporal features could be extracted using RNN, while the spatial features could be extracted using CNN. Previous research discussed the potential of CNN-LSTM to predict PM2.5. However, this prediction is only based on 1-dimensional data so that the spatial relationship of the sensor networks cannot be analyzed [27]. The model we designed is capable of recording a 2-dimensional dataset with a feature extraction that could record spatiotemporal patterns. Furthermore, in order to extract high complexity features, a large number of neuron networks are required which are arranged in stratified layers. Our proposed model uses deep learning by combining CNN and RNN into several layer neurons. Both approaches layers are combined to process spatiotemporal data to provide better predictions than the existing approaches.

## 3. Methods

Here, we describe general knowledge about both convolutional and recursive NN. RNN is a feedback neural network that is used to extract patterns in a sequential dataset. Meanwhile, CNN is a type of NN which is its input calculated by convolution to extract spatial patterns. At the beginning of this chapter, several types of recursive networks were introduced, *i.e.* RNN, GRU, and LSTM. In the end, we introduce CNN and convolution RNN which able to learn more complex datasets.

### 3.1. Recursive/Recurrent Neural Network (RNN)

RNN is a form of Artificial Neural Networks (ANN) architecture that is specifically designed to learn sequential data. It is usually used to process tasks related to time series data. RNN is quite widely used in solving sequential problems [28], *e.g.* Natural Language Processing (NLP), speech recognition, machine translation, video classification, stock prediction, and weather forecasting. The idea behind the RNN architecture is how to exploit sequential data structures. RNN is a modification of NN with the output fed back to the next NN input repeatedly. It means that the same operation is performed for each sequence element, with the output depending on the current input and the previous operation. In essence, RNN focuses on the nature of the data where the previous $x_{(t-1,t-2,...,t-T)}$ and present time $x_t$ affects the next time variable $x_{t+1}$ . RNN does not simply throw information from the past in the learning process. It is what distinguishes RNN from ordinary NN. The way an RNN could store information from the past is by looping through its architecture, which automatically stores information from the past into distributed weights.

### 3.2. Gated Recurrent Unit (GRU)

The idea of designing a GRU network is each iterative unit captures dependencies in different time scales adaptively [29]. As an analogy, humans don't need to use all the information in the past to be able to make decisions now. For example, it could be analogized to a rainfall prediction system in an area with two seasons, *i.e.* dry season and rainy season. Information from the past about rainfall in the dry season will not contribute significantly to decision making when the current conditions are in the rainy season. Inside GRU, the information flow control component is called a gate and the GRU has 2 gates, namely a reset gate and an update gate. The reset gate determines how to combine the new input with past information. Meanwhile, the update gates determine how much past information should be stored while reading/generating a sequence.

### 3.3. Long Short-Term Memory (LSTM)

LSTM is a type of RNN with the addition of a memory cell that could store information for a long time. LSTM is proposed as a solution to overcome the vanishing gradient in RNN when processing long sequential data. LSTMs are able to learn long-term dependencies that were previously a weakness in RNN. LSTMs also have repeating connections or chain-like structures. The difference between LSTM and RNN lies in the layers contained in each LSTM cell. In each LSTM cell, there are three Sigmoid functions and one Hyperbolic Tan function. For long-term dependency problems, LSTM could handle noise, distributed representation, and continuous values [30].

### 3.4. Convolutional Neural Network (CNN)

CNN is a type of NN with its inputs in the form of two-dimensional data so that the linear operations and weight parameters on CNN are different from NN. Inside CNN, linear operations use convolutional operations, while weight is no longer one-dimensional, but in four dimensions which is a collection of convolutional kernels. CNN is inspired by the Visual Cortex, which is the part of the brain that processes information in visual form. With such an architecture, CNN could be trained to understand the details of 2-dimensional data even better. That way, CNN could capture spatial and temporal dependencies in an image when operated with relevant filters so that it could be used to predict a short sequence of 2-dimensional data [31, 32]. To provide a better understanding of a complex pattern, CNN is usually designed with many interconnected layers, so it is commonly called a deep learning architecture.

### 3.5. Convolutional Recursive Neural Network (CRNN)

The traditional convolutional layer extracts features from the data by applying non-linearity to the activation function of the input. CRNN improves this feature extraction process especially for the case of sequential data, by entering data into the RNN and using the output of the repeating unit to compute the extracted features. This architecture exploits the fact that a window containing multiple frames of sequential data is a temporal feature that might encapsulate valuable information. Meanwhile, spatial features are recorded in the convolution layer which is directly connected to the input layer. CRNN model could be used to predict the dataset with long sequence format, *e.g.* Internet sentiment analysis [33] and weather forecasts [25].

## 4. Design of Predictor Model Based on CRNN

Here, we describe in detail the basic concepts of designing a predictor model on a massive scale sensor network. The initial section describes the process of data collection and preprocessing. Next, we explain a simple forecasting model that is built from layers of simple RNN. We also introduce the multi-input RNN approach by taking into account the measurement results of several nearby nodes. Finally, we describe our approach that utilized a deep learning model by using CRNN to extract spatiotemporal features in the PM2.5 dataset.

### 4.1. Tools and Dataset

The dataset is obtained from sensor nodes as part of the ‘Airbox’ sensor network that provides real-time PM2.5 monitoring services. The system has been implemented throughout Taiwan and includes thousands of community-installed sensor nodes and hundreds of government-owned nodes. This air quality monitoring system is hereinafter referred to as the ‘Airbox’. As part of a transformative computing system, the scheme we propose is shown in Figure 1. This system does not completely change the already implemented 'Airbox' system. This schema modifies the server node by implementing AI technology for a more precise forecasting model. The specifications we use in this research are shown in Table 1. As a limitation, we use several of them *i.e.* 268 sensor nodes scattered across Central Taiwan. We use data in the area to describe a larger forecasting system for the future. Central Taiwan is suitable as a research case because as referred to [1], this area reflects the condition of Taiwan as a whole. It is also home to the third-largest coal-fired power plant in the world, which produces large emissions of carbon and micro particles.

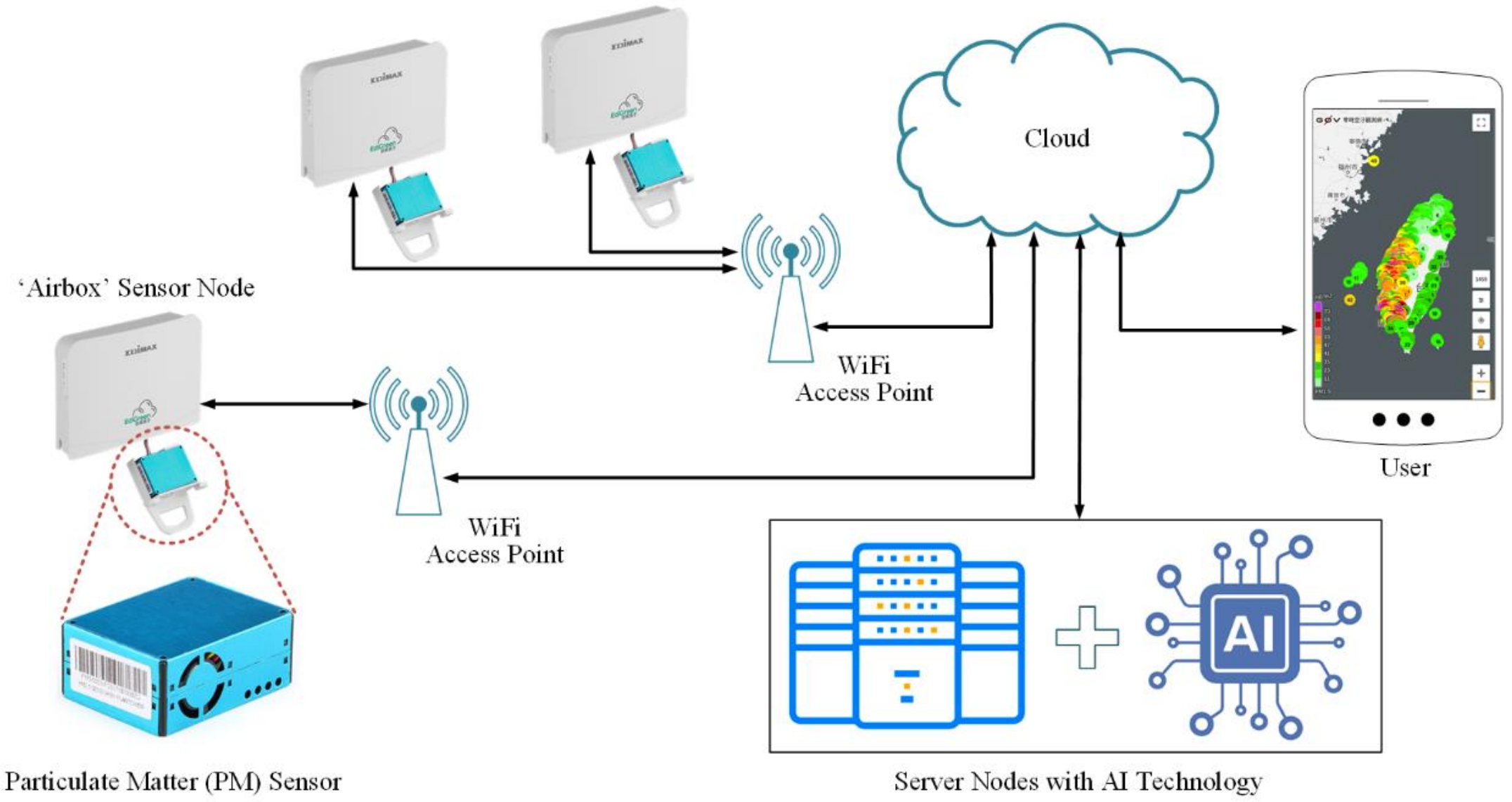

Figure 1: Schema of transformative computing utilized WSN that served by PM2.5 sensor nodes named ‘Airbox’ combined with AI cloud computing.

Table 1. The specification of sensor nodes and server node used in this research.

| Sensor Node | |
|---|---|
| Size | : 148.4mm x 111.5mm x 45mm |
| Connection | : Wi-Fi IEEE 802.11b/g/n |
| Temperature sensing range | : 0~60°C, Accuracy: ± 1 °C |
| Humidity sensing range | : 0~100%RH, Accuracy: ± 5% RH |
| PM2.5 measurement range | : minimum 0.3µm |
| PM2.5 measurement efficiency | : 50% @ 0.3um, 98% @ >= 0.5 um |
| Power supply | : Micro USB port x 1 DC 5V |

| Server Node | |
|---|---|
| Processor | : Dual 20-Core Intel Xeon E5-2698 v4 2.2 GHz |
| RAM | : 256 GB |
| GPU | : NVIDIA Tesla P100 (3584 CUDA Cores) |
| GPU memory | : 32GB HBM2 |
| OS | : Ubuntu 64-bit |
| Environment | : Python with Keras (TensorFlow backend) |
| Library | : numpy, pandas, pyplot, and folium |

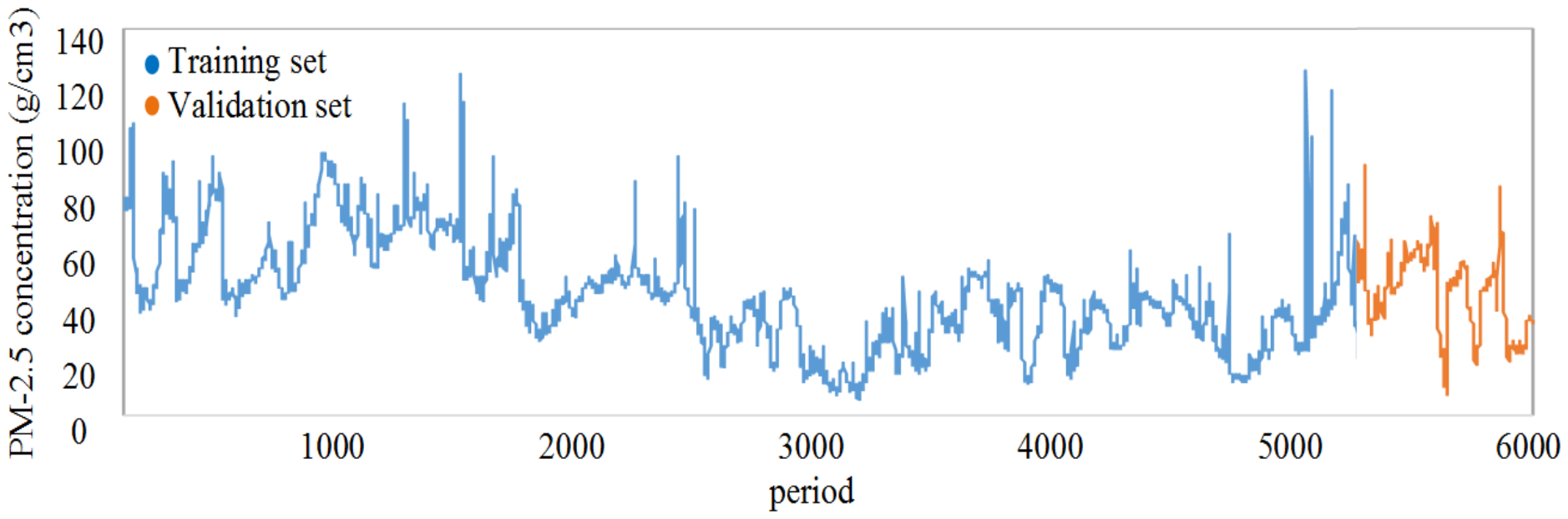

Figure 2. A single dataset from 1 sensor node measured in 1 month with 5-minutes sampling time, consists of 90% as training data and 10% as testing data.

In visually, the sensor placement in this area is uneven with higher density nodes in urban areas. Meanwhile, remote areas receive less sensor distribution. It means the density of data obtained from urban areas will be higher. Data resolution in urban areas is better than in rural areas. However, the distribution of sensors in urban areas is not very uniform. It is a challenge how to design data that is, in the beginning, are in non-uniform positions so that it could be processed by deep learning while maintaining spatiotemporal data uniformity. Preprocessing of data needs to be done so that the position of data placement could be maintained so that the correlation between data is guaranteed. It means that the dataset that is close in term of recording locations have a stronger relation than data that is farther away.

PM2.5 dataset was collected for one month *i.e.* in September 2020 with a sampling rate every 2 hours. We perform experiments on a single dataset and a complete dataset covering all sensor nodes. On a single dataset, we applied a simple RNN to measure the best performance of the three types of recursive layers, *i.e.* simple RNN, GRU, and LSTM. While the CRNN model is performed to extract the hidden spatiotemporal features on the full dataset. Figure 2 is a single dataset from a sensor node during one month of measurement. Visually it could be seen that the data has a rather random pattern but still has a repetition pattern every day. Although this pattern is not very visible, a multilayer RNN network could be used to extract the pattern. In general, the learning process begins by using a number of repeated sequences from the dataset. From each sensor dataset, we use 90% as training data and 10% as testing data. The evaluation process is carried out using training losses, accuracy, and RMSE which is measured using the predicted results as input.

## 4.2. Design of A Simple Predictor Using RNN layers

The design of the RNN model is carried out using the python programming language with the support of the TensorFlow library. In this section, we want to see how each type of recursive layer generates different performance. A dataset of a single sensor was used as experimental material. We present the lowest losses and the best accuracy generated by the three types of recursive layers *i.e.* RNN, GRU, and LSTM. The approach that has been implemented is by using a hybrid neural network [34] and multivariate LSTM [35]. This approach does not involve the extraction of spatiotemporal features. We compare several approaches with our model framework, in terms of greater precision, accuracy, and scalability.

At first, we determined the target of this forecasting system, which is forecasting with a sampling time of 2 hours for the next 24 hours with the past 1 days dataset. Each day will generate 12 data in a sequence. Then, we define a model specification *i.e.* input, the number of neurons per layer, and output. In accordance with the target, the number of input neurons is determined to be 12 x 2 days = 24 neurons. A compilation process is carried out to arrange these layers into a deep learning model. Furthermore, the process of arranging the layers into a model is carried out sequentially with each type of recursive layer having a different number of layers varying from 1 to 5. Between each layer, a dropout layer is inserted to prevent overfitting and also accelerate the learning process.

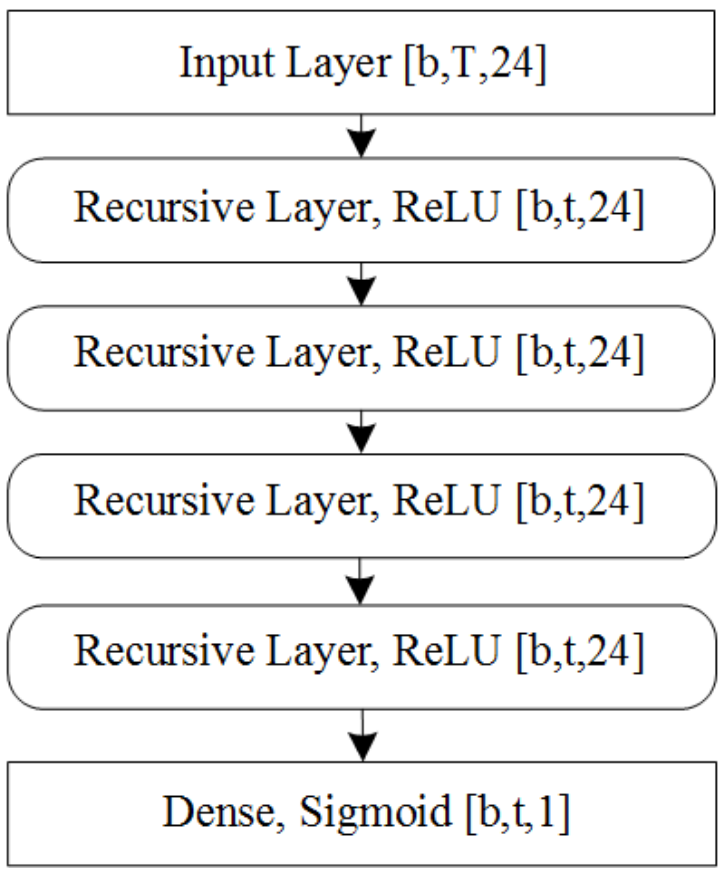

Figure 3. Design of a simple predictor using several recursive layers *i.e.* RNN, GRU, and LSTM.

Table 2. The different layer depth variations on a simple RNN affect the execution time without significantly increasing accuracy.

| Number of Layers | Number of Parameters | Training Time (s) | Losses | Accuracy |
|---|---|---|---|---|
| 1 | 6,49 | 11 | 0.0143 | 0.7583 |
| 2 | 1,825 | 19 | 0.0084 | 0.8943 |
| 3 | 3,001 | 35 | 0.0011 | 0.9514 |
| 4 | 4,177 | 42 | 0.0025 | 0.9011 |
| 5 | 5,353 | 67 | 0.0012 | 0.9102 |

Table 3. The difference between a simple RNN, GRU, and LSTM by using different number of input channels.

| Recursive Layer | Number of Parameters | Input Channels | Losses | Accuracy |
|---|---|---|---|---|
| Simple RNN | 4,177 | 1 | 0.0020 | 0.9156 |
| | 4,201 | 2 | 0.0018 | 0.9211 |
| | 4,225 | 3 | 0.0019 | 0.9180 |
| | 4,249 | 4 | 0.0025 | 0.9011 |
| GRU | 12,769 | 1 | 0.0015 | 0.9455 |
| | 12,841 | 2 | 0.0014 | 0.9493 |
| | 12,913 | 3 | 0.0018 | 0.9330 |
| | 12,985 | 4 | 0.0020 | 0.9190 |
| LSTM | 16,633 | 1 | 0.0011 | 0.9514 |
| | 16,729 | 2 | 0.0007 | 0.9673 |
| | 16,825 | 3 | 0.0015 | 0.9404 |
| | 16,921 | 4 | 0.0019 | 0.9211 |

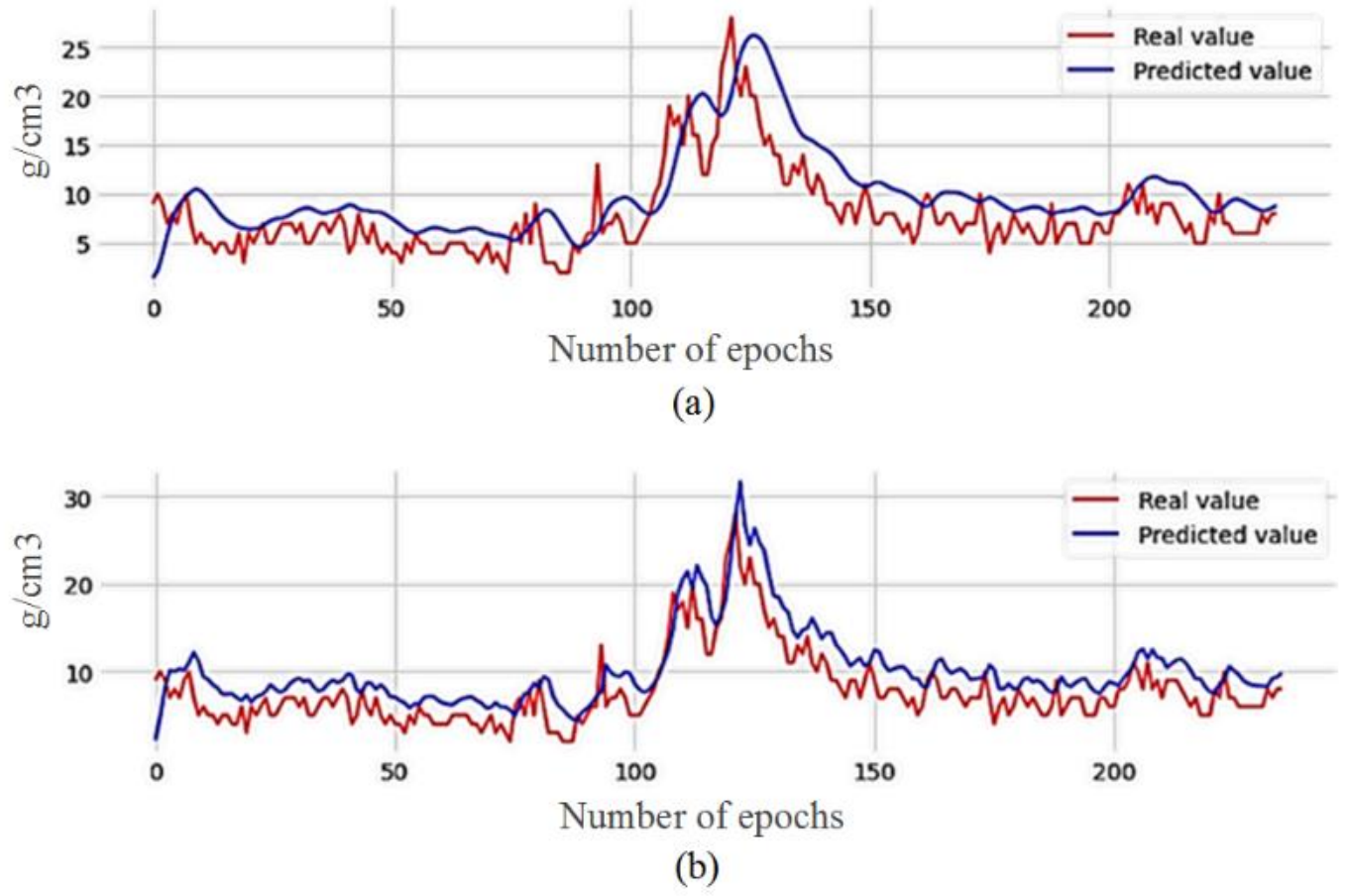

Figure 4. The results performed by using a recursive model *i.e.* (a) GRU and (b) LSTM without involving the result as the next input.

From Table 2, we could conclude that the 4 layer configuration provides optimal performance without sacrificing a lot of neurons. The use of more than 4 layers does not provide significant performance, thus decreasing computational efficiency. We use this 4 layer configuration to build our model.

To see the best performance on the forecasting system, we also take into account the measurement results of the closest nodes around the test node. In theory, the more data that is included in the calculation, it affects the accuracy of the system. Therefore, each input has a channel variation from 1, 2, 3, and 4 which represents the number of closest nodes that are used as modeling references. The output is 1 neuron which represents the next 2-hours prediction. Furthermore, because the network used is a recursive type, the input will be shifted and the output will represent the predicted results of the next hour. After compilation, the number of parameters in the LSTM and GRU network are 4 and 3 times more than that of on the simple RNN as shown in Table 3. The greater the number of parameters, the longer the training process will take. Likewise, the greater the number of layers, the greater the number of parameters. With the increasing number of parameters, the more local and global features that could be stored on it. It could be seen that the best accuracy is performed by LSTM. Even if we include nodes around the location into the learning model, the performance does not improve significantly. Increasing the number of input channels actually increases the number of neurons without a significant contribution to accuracy. The prediction results look good visually and are able to replicate real-world value shown in Figure 4. However, these results are obtained for forecasting one step ahead. To predict the next few steps by involving the results as inputs, this 1-dimensional recursive model has not provided good performance.

### 4.3. Design of Predictor Framework Using CRNN Layers

The design we are proposing consists of 2 main parts *i.e.* preprocessing and CRNN modules which are composed of CNN encoder, RNN, and CNN decoder, shown in Figure 5. The preprocessing module translates the all records from the total 268 sensor nodes into a 4-dimensional sequence data in the form of heat maps. The data is arranged in relatively to other data based on spatial and temporal relationship. Then, the arrangement of each layer of the proposed model is explained in detail. Finally, the learning strategy is described for producing measurable predictions.

#### 4.3.1. Preprocessing

Preprocessing converts raw data with an unequal distribution into structured data, which has strong tied in spatially and temporally. This preprocessing involves the geographical position of each node and its measurement time. That is, each data is collected and arranged based on both the time in sequence as well as the spatial information in geographic location according to Voronoi diagram shown in Figure 6, where Voronoi diagram [36] is a mathematical method to partition a plane into regions closing to each of a given set of objects. The data is collected by the node named ‘Airbox’ in this research, which the PM2.5 data is detected and generated by using five minutes sampling rate. We resample the data every 2 hours for the last 2 days (24 sequences) as a training set, while 12 hours of subsequent data were used as ground truth. The dataset is processed by the model, then the results are the next 24 hours prediction values. In theory, the more time to predict, the more relative error if we include the current result as the next input. For considering the concerns mentioned above, we compare the approach that has been taken temporally with the deep learning model emphasizing the spatiotemporal relationship in the dataset.

The data are not evenly distributed spatially. Unevenly distributed data is difficult to feed directly to the deep learning model. Therefore, we split the data spatially into several sectors H. Taiwan has a main island that stretches from North to South. For the experiment, we divide Taiwan into small sectors with a resolution of 40 rows *m* and 40 columns *n* starting from coordinates (23.90, 120.37) to coordinates (24.45, 121.020). We choose this resolution so that one sector only covers 1 to 10 nodes. If the resolution is higher, the number of neurons used will be bigger which will affect the computer's ability to complete the training process. This resolution is the most optimal in our understanding.

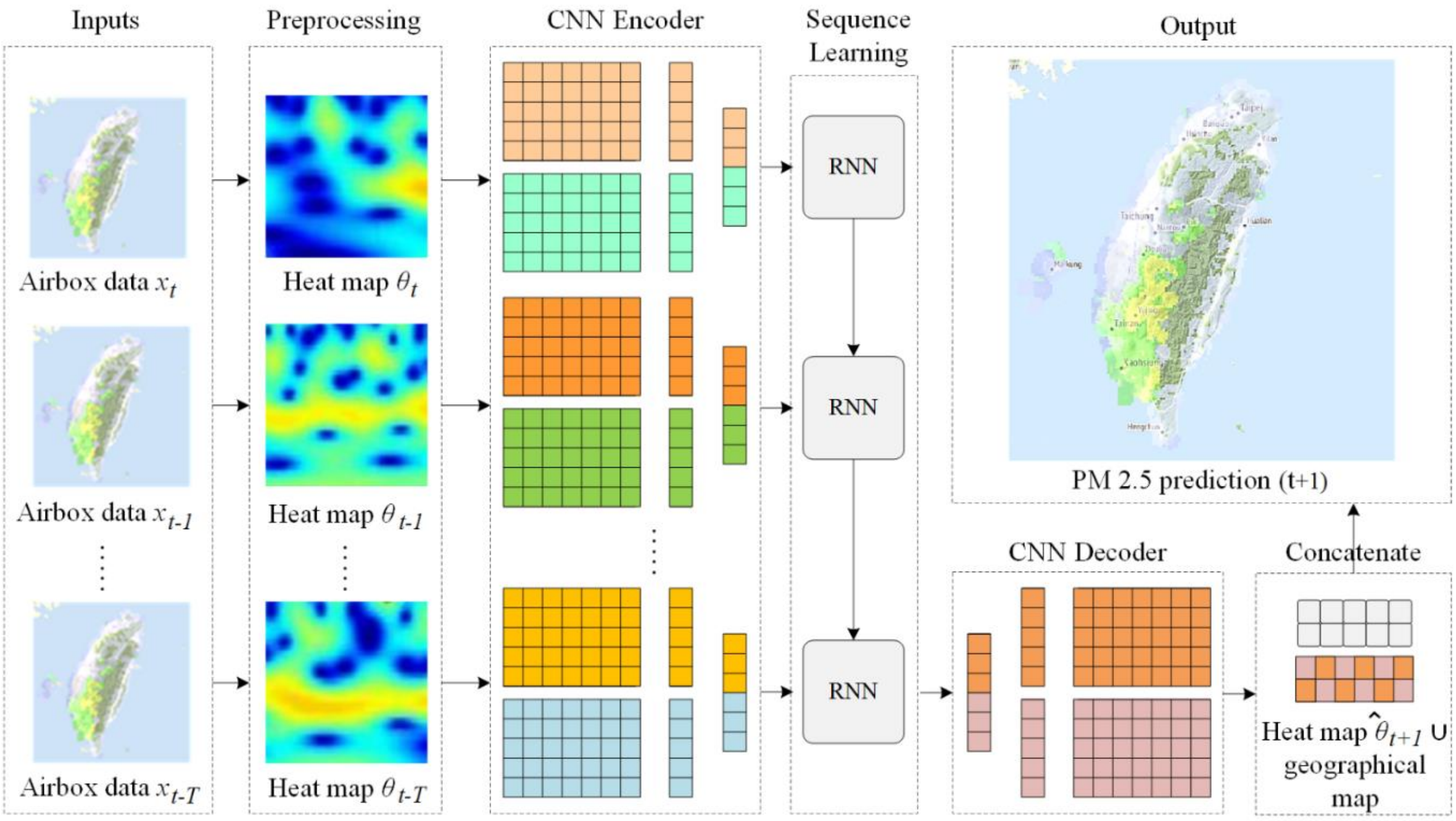

Figure 5. Design of Predictor Framework based on CNN-RNN, consist of input, preprocessing, CNN encoder, RNN-based learning, and CNN decoder.

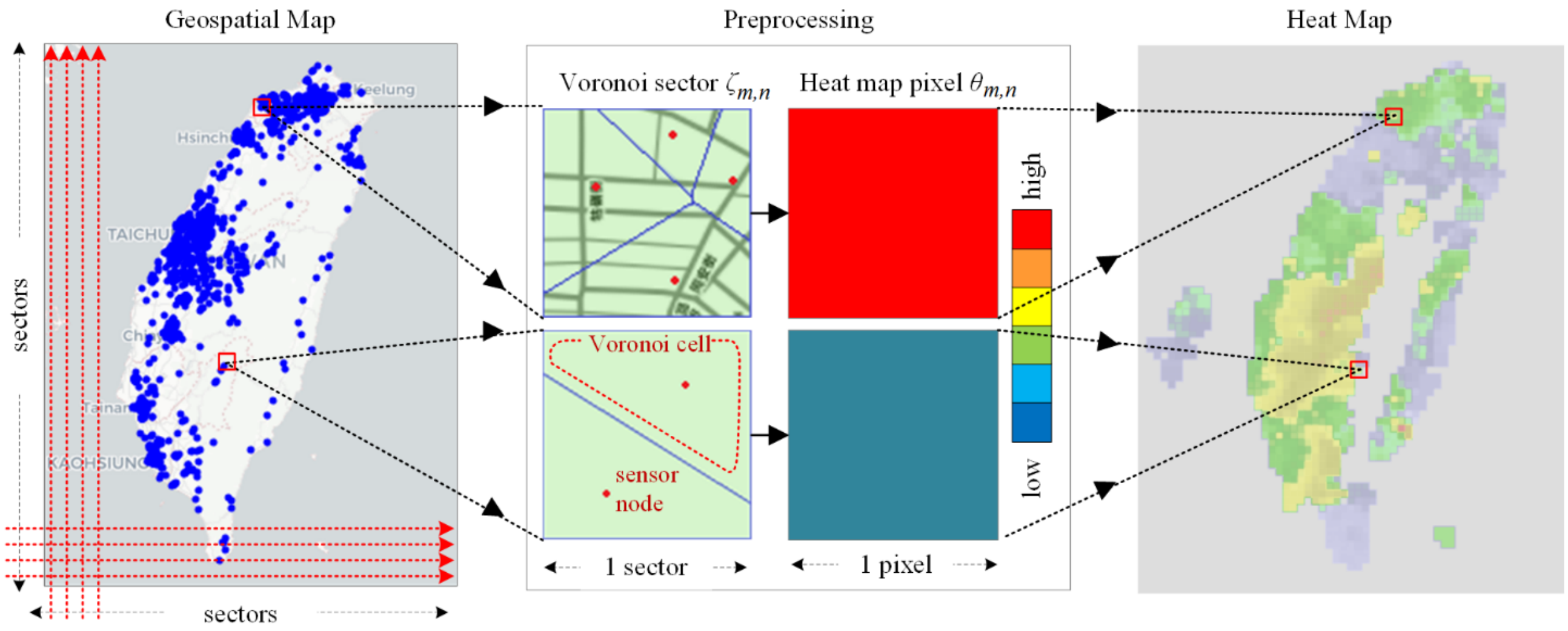

Figure 6. The preprocessing procedure involves the conversion process which transfers the geospatial position of sensor nodes into Voronoi map. Then, the heat map is generated by using mean value operation for all nodes in each Voronoi section.

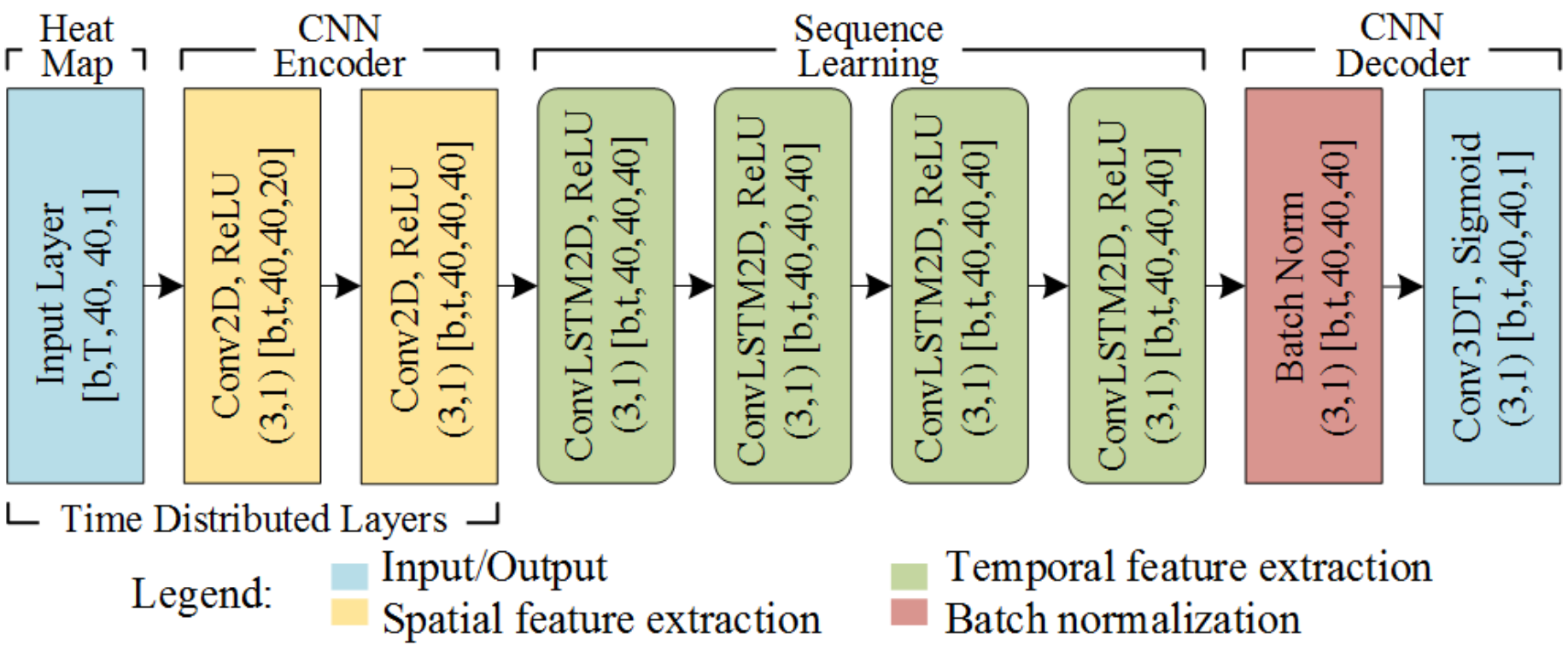

Figure 7. A layer-wise schematic of our deep learning model. It exploits autoencoder-like with CRNN strategy.

Each sector provides similar features for several sensor nodes. In one sector, it is consists of several cells that represent the coverage area of each node as shown in Figure 5. Each sector is connected to 1 input neuron. Therefore, Equation (1) is needed to convert the values of several sensor node measurements into an aggregation value that represents the value of the sector. We use the Voronoi diagram to calculate the contribution of a node in a sector. A Voronoi diagram is a division of the area of a plane into sections based on the distance from points on a specific subset of the area of the plane. The diagram divides a sector based on the position of the nodes in the plane into several Voronoi cells with a certain area. Each sector has a certain value and it forms a heat map value which represents the PM2.5 pollution level. Given sensor nodes in geographical position $(i, j)$ inside certain sector $\zeta_{m,n}$. The following equation is used to compute the heat maps which will be used as input to the deep learning model on the next process.

$$\theta_{m,n} = \begin{cases} \sum x_{i,j} \dfrac{v_{i,j}}{V_{m,n}} ; & x_{i,j} \in \zeta_{m,n} \\ 0 ; & x_{i,j} \notin \zeta_{m,n} \end{cases} , \tag{1}$$

where $\theta, x, v, V$ are heat map value, PM2.5 value, volume of Voronoi cell, and volume of sector respectively. In addition, heat map is a useful tool to see the activity of the research object with color indicators. This color represents how active, much, and intensely an object behaves in a plane. We illustrate the PM2.5 propagation rate over time with the descriptions on this heat map.

### 4.3.2. Going Deeper into CRNN Model

We use a bottom-up problem implementation strategy to design the prediction model. The research that has been conducted uses a recursive learning base to deal with forecasting atmospheric variables. We were inspired to complete sequence learning using the LSTM module, with considerations of better performance than other recursive layers as shown in section 4.2. The proposed model overcomes the problem of lack of heat map depiction by capturing short and long-short spatiotemporal cues at local and global levels via 3D convolution and LSTM modules.

The proposed model is an improved version of the simple recursive network in Figure 3. We start from the basic model as shown in Figure 5. This model is inspired by autoencoder, but it uses a Conv-LSTM2D layer instead of the standard convolutional layer. We upgraded the model to the CNN-LSTM model. Each spatial data from sensor nodes will be processed at one time using the CNN module. The CNN is applied recursively to all inputs $\theta_{m,n}(t, t-1, t-2, \ldots, t-n)$, *i.e.* the heat map of PM2.5 concentration. Then, each spatial feature will be processed using the ConvLSTM2D module to extract temporal features from the data slice sequences. We tested both models on the selected dataset on the 'Airbox' records that were recorded for a month. Then based on empirical evidence we completed the model and conducted extensive experiments on 24 forecast sequences compared with the ground truth dataset.

Figure 7 shows the CRNN network which illustrates in detail our proposed model. The $k$ kernel size, step rate $s$, and output dimensions are expressed in the following order and include brackets $(k, s)$ $[b, t, M, N, D]$ on each layer. Where, $b, t, M, N,$ and $D$ represent the batch size, the number of samples taken by the Conv-LSTM module to capture temporal information, the height and width of the frame, and the number of output feature maps. The network has 9 layers with 78,261 trainable parameters that integrate four main components, *i.e.* CNN encoder, recursive layer, CNN decoder, and recursive predictor. It maintains a constant number of filters (16) in each layer (except the penultimate layer, which yields 20 feature maps) and the kernel size $k = 3$. Therefore, the spatial dimensions of the feature map are maintained similar from input to output. Thus, the encoder final layer generates a feature map that has a spatial dimension of 40 × 40. Because the network input layer accepts a frame with 40 x 40 spatial dimensions, so the preprocessing data should be done before. To achieve a precisely decoded feature map, there are four sequentially connected mini-decoder blocks. Where, each block group Conv2D and ConvLSTM2D, sequences. Therefore, the last layer of the decoder generates a feature map with the same spatial dimensions as the network input. The final classifier module consists of Batch Normalization (BN) and 3D Convolutional layer with Sigmoid function as the classifier. Furthermore, this map feature is merged with the raw geographical map using the concatenate operator. Thus, the output of this model is a probability map of the next frame which is estimated based on the $T$ observed frames.

### 4.3.3. Training Strategy

Experiments were carried out on the sequence of the dataset 24 hours back to determine the forecast results for the next 24 hours. This means, the number of forecasts compared with the reference dataset is 1: 1. We keep this ratio not too small so that the number of neurons required in the modeling is not too large. While the ratio is also not too close to 0 so that the model has sufficient references to produce good forecasting accuracy. This approach is more precise than the random selection of frames to solve the data sequence prediction case. Python with Keras (TensorFlow backend) is used as the supporting software for this research. The network trained on NVidia Tesla 32 GB working in desktop DGX-1. On average, the training takes about 20 minutes to 30 minutes depending on the order properties of the dataset. The model is trained individually on each data set with the Adadelta optimizer described in Equation (2). Learning rate is set to 0.0002 with a scheduler reducing learning speed by factor of 0.8.

$$E = \frac{-1}{n} \sum_{n=1}^{N} \left[ p_n \log p_n + (1 - p_n) \log(1 - p_n) \right], \tag{2}$$

where it takes two inputs; first one is the output from the final layer of the network with dimension of $b \times T \times M \times N \times D$, which maps the pixel probabilities $p^n = \gamma(x_n) \in [0,1]$ using Sigmoid classifier $\gamma$.

## 5. Results

This experiment is carried out by involving the training and validation dataset. To produce optimal measurements, it is necessary to do the tuning process on the model that has been made. Then the test results will be presented, followed by a comparative analysis of several approaches and their effectiveness in generating a PM2.5 prediction map. Finally, we provide the result into a sequence map compared with the conventional approach.

### *5.1. Fine Tuning the model*

We tuned several parameters of this CRNN model, consisting of the number of neurons per layer, epoch, batch size, and delay time. The learning and testing process is carried out on our server. The tuning process needs to be carried out to produce the smallest number of parameters so that the training process could be run in the shortest possible time. In the tuning process, it should be noted how much the optimal value of the parameter number of neurons, epochs, and batches. The more the number of neurons, the slower the training process and many neurons that do not contribute too much to the training. Too many epochs result in the length of the training process even though the accuracy does not improve. This also applies if the number of batches given is too large. From repeated experiments to test the most optimal parameters of the model, we use batch size = 20 and epoch = 500. We tested 5 models *i.e.* NN [34], LSTM [25], CNN [31], Convolusional+LSTM (ConvLSTM) [32], and our proposed CRNN model with a total parameter of 8,459, 16,633, 43,861, 46,281, and 78,261 respectively.

### 5.2. Evaluation

We measure standard performance based on Root Mean Square Error (RMSE) which evaluates the similarity between the heat map predictions $\hat{\theta}_{t+1}$ and the ground truth $\theta_{t+1}$ . This standard is a measure of the average relative error for each pixel. The lowest the ratio, the better the performance of the prediction system. Let $\theta = \theta_{1,1}, \theta_{1,2}, \theta_{1,3}, \ldots, \theta_{m,n}$ be the ground truth data and $\hat{\theta} = \hat{\theta}_{1,1}, \hat{\theta}_{1,2}, \hat{\theta}_{1,3}, \ldots, \hat{\theta}_{m,n}$ be the predicted heat map. Then, the RMSE could be defined as:

$$NRMSE = \frac{1}{\bar{\theta}} \sqrt{\frac{1}{MN} \sum_{m}^{M} \sum_{n=1}^{N} \left(\theta_{m,n} - \hat{\theta}_{m,n}\right)^2} , \qquad (3)$$

where $\bar{\theta}$ , $M$ and $N$ are mean, max height, and max weight of the original heat map, respectively.

We also test how durable the models when implemented with real conditions where there are possibilities of errors in the data measurement process. In this condition, with the increasing number of nodes connected in the WSN, the chances of sensor damage, data transmission errors, and even nodes that do not transmit data could occur. Because of that, we evaluate the robustness of the model by adding noise with the standard deviation σ. The greater the σ, the greater the damage to the dataset. An approach that utilizes spatiotemporal feature extraction could be an alternative to reduce the dataset fault because of noise.

### 5.3. Results

The test results are divided into two, namely qualitative and quantitative assessments. Quantitative assessments are obtained using the NRMSE graph as shown in Figures 8 and 9. Meanwhile, quantitative assessments are carried out by visually examining the prediction results with ground truth as the target described in Figure 9.

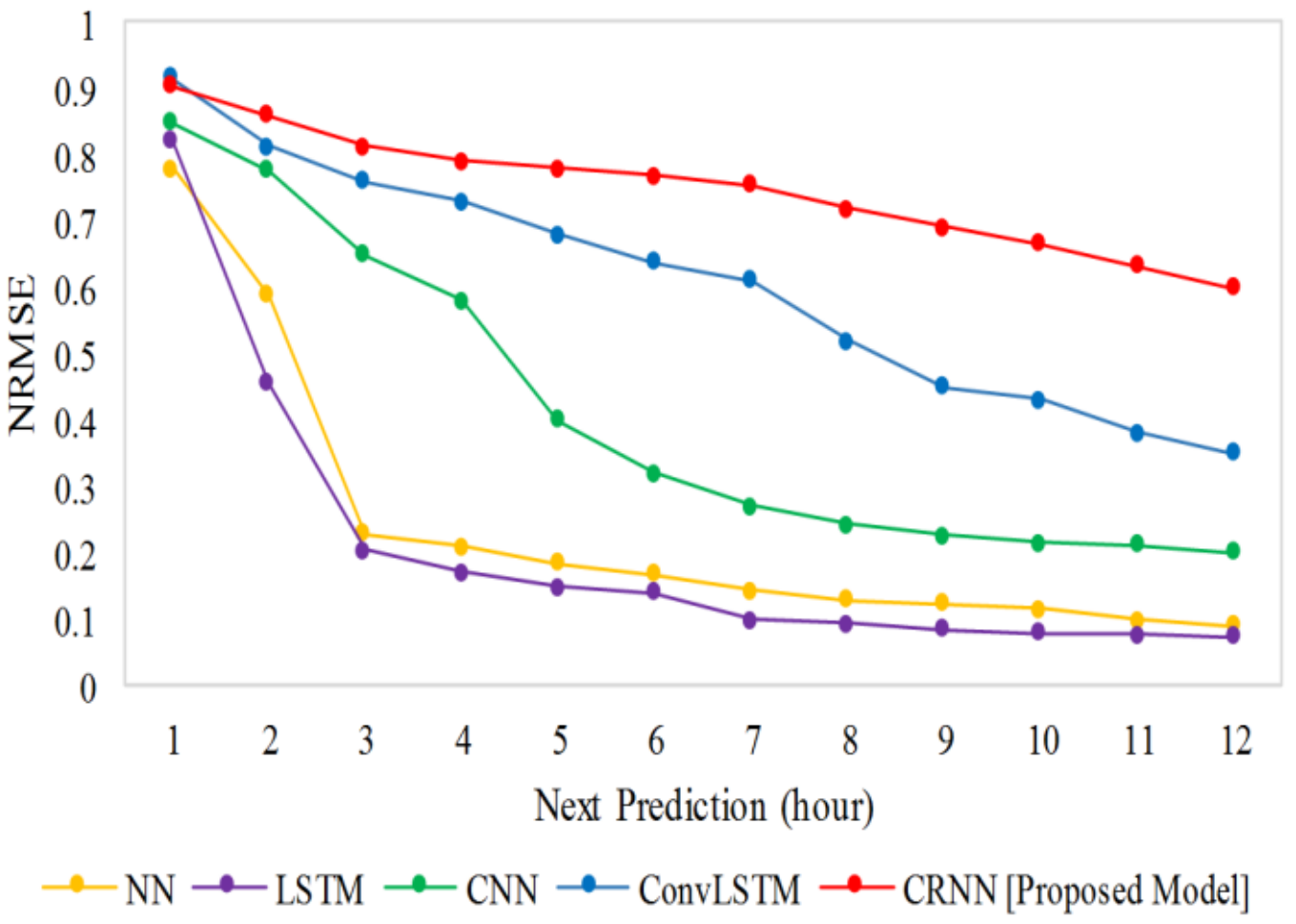

Figure 8. The RMSE of 5 predictive models involving the use of results $\hat{\theta}_{t+1}$ as input $\theta_t$

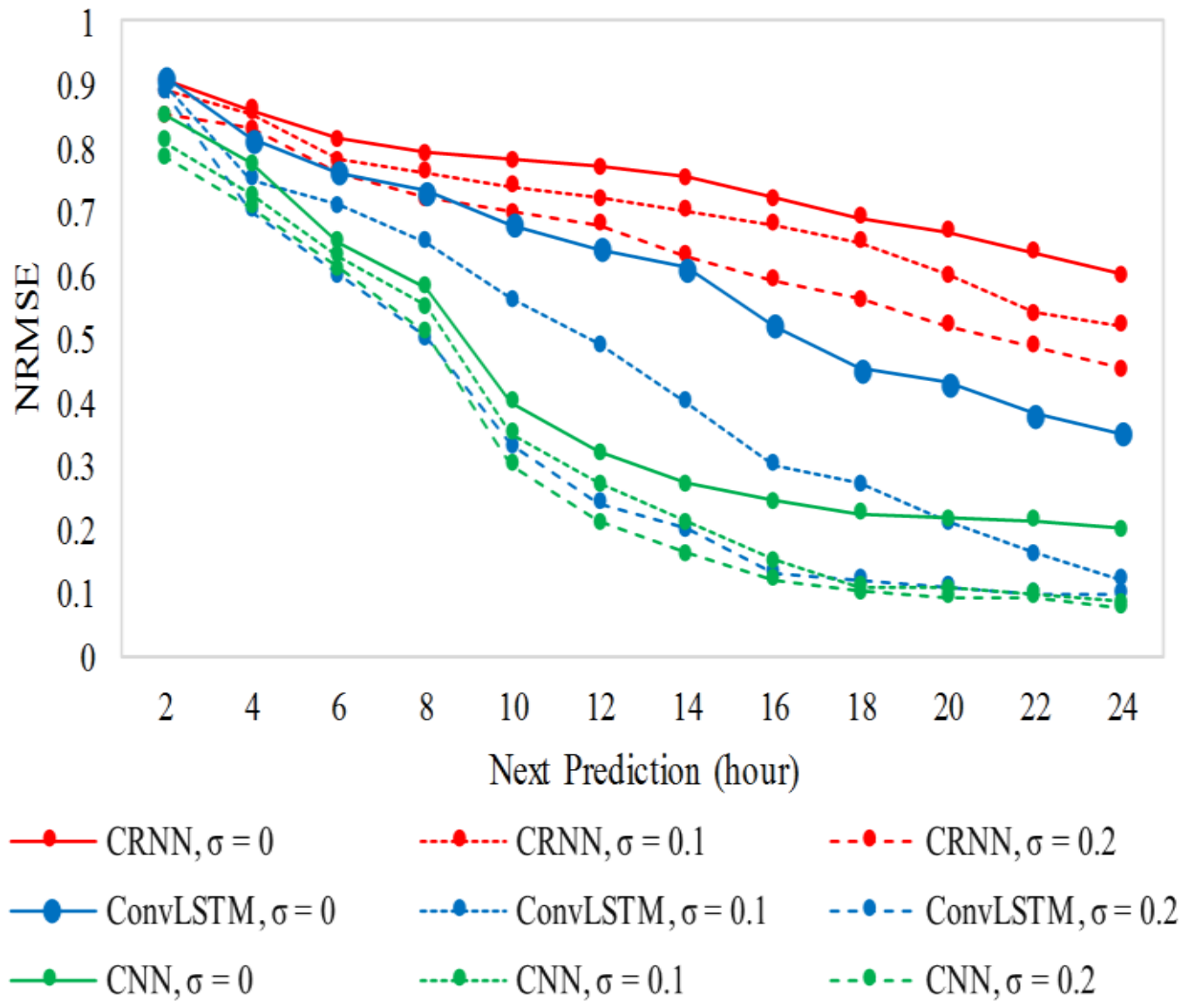

Figure 9. RMSE of the top-3 models *i.e.* CRNN, ConvLSTM, and RNN with several of noise addition $\sigma$ .

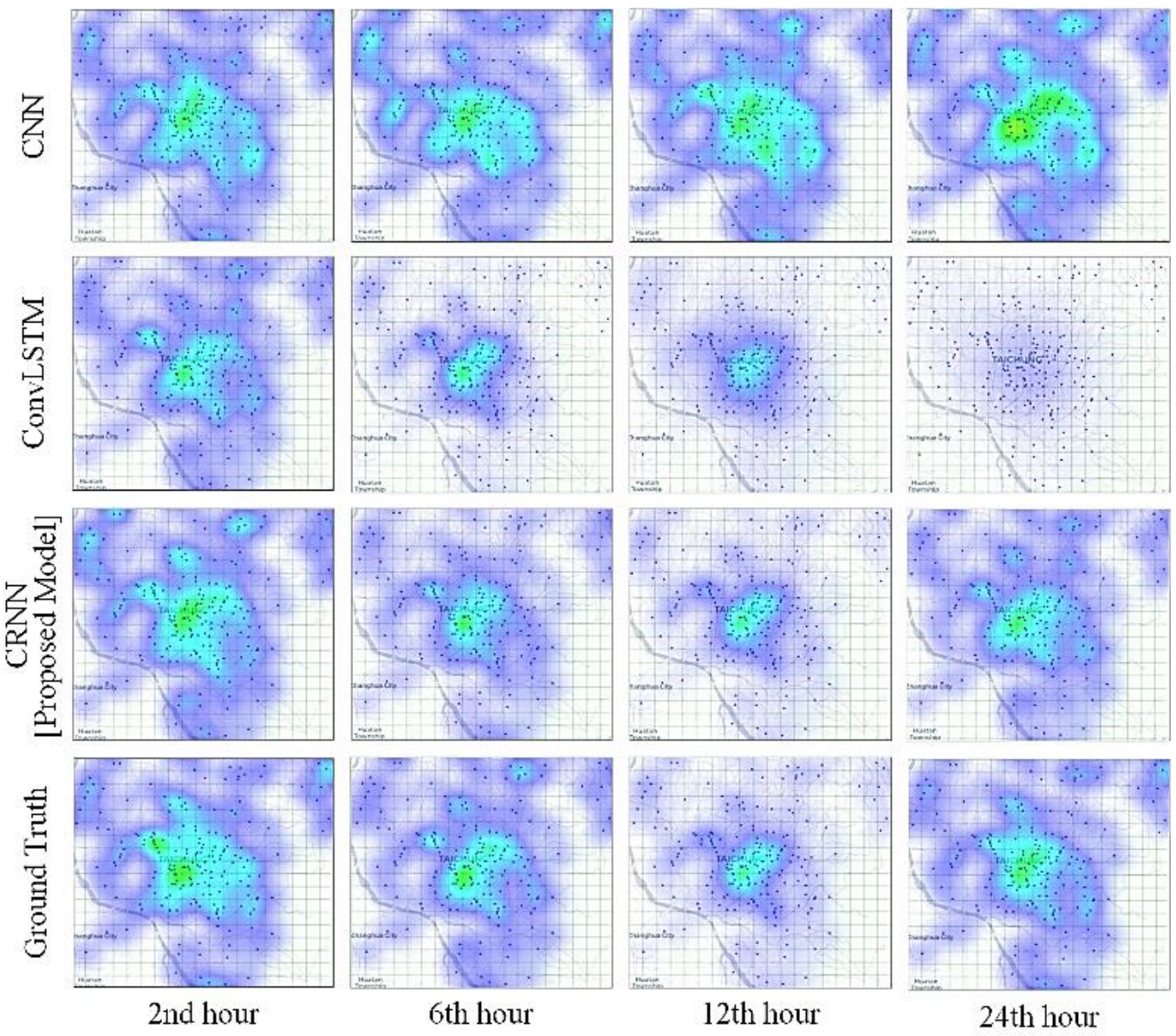

Figure 10. Propagation map of PM2.5 in Central Taiwan generated by several deep learning models.

As provided in Figure 8, all models perform good prediction proven by an accuracy value above 0.75 for the period $t+1$. In this period, ConvLSTM generates the highest accuracy results with an NRMSE value of 0.912 followed by our proposed model with a value of 0.908. The other models generate NRMSE values of 0.87, 0.85, and 0.78 for CNN, LSTM, and NN, respectively. 1-dimensional forecasting methods such as LSTM and NN are not able to keep up with the performance of 2-dimensional forecasting models, *i.e.* CNN, ConvLSTM, and CRNN. Furthermore, the next step for the prediction $t+2$ shows a significant performance difference between the five models tested. The 1-dimensional forecasting model cannot compensate for the other three models with a significant decrease in the NRMSE value. This could occur because the weight distribution in the 1-dimensional network model is very small and there are no long-term memory modules. The spatiotemporal data pattern is very suitable to be solved with a 2-dimensional forecasting model.

In Figure 9, it could be seen that the proposed model is able to produce the best performance compared with other 2-dimensional models when dealing with noise that occurs in the measurement system. Even when given noise $\sigma = 0.2$, our model is still able to compensate for the performance of the ConvLSTM model without noise addition. The addition of noise actually worsened the performance of the other 2 models, so it is evident that our proposed model has an advantage over the previous approach. Our model is capable to extract spatiotemporal dataset because it has the two separate modules. One is CNN which is used to extract spatial features, and the other one is LSTM which is used to extract temporal features.

## 6. Discussions

From this study, it can be analyzed that the LSTM model is generally better than the GRU and RNN models in overcoming the high randomness level of the data on PM2.5 data. Furthermore, this model could be used to predict the direction and rate of PM2.5 propagation. The more nodes involved, the better the forecasting performance. On a simple LSTM model, the use of more than 2 nodes actually makes performance decrease. Based on our tests, using 1 input on the LSTM module generates the best performance without sacrificing an unnecessary increase in the number of neurons. It happens because the complexity of forecasting is higher with the increasing number of variables involved. A simple recursive network has not capable to extract these patterns. Moreover, the conventional model could only capture temporal patterns. In fact, forecasting that involves massive datasets should have spatial patterns that are sometimes not recorded when using only recursive models.

Our proposed model provides an advantage in areas where sensor distribution is sparse. The model generates a higher resolution with a more balanced distribution by taking into account the training data from other nodes. Forecasting resolution also increases because a node inside the heat map has a spatial strong relation to the surrounding nodes. The spatiotemporal features could be recorded better with our proposed approach. By adding datasets throughout the year, local and global features could be extracted precisely. The period of forecasting could also be extended if the dataset is larger. In addition, with the increasing number of training datasets, it needs to be accommodated by an increase in the number of neurons and the depth of the layer. Increasing the number of neurons will increase the training time. However, the use of CRNN has been proven to be used in PM2.5 forecasting system applications without considering other atmospheric variables. The influence of other atmospheric variables could be suppressed by learning the propagation of PM2.5 pollutants using a deep learning model so that it improves the efficiency of data processing in a massive-scale WSN.

## 7. Conclusions

This paper presents the model according to the idea of transformative computing, which deals with the collected real-time information outputted by a large-scale sensor network by using a deep learning CRNN model in order to predict PM2.5 concentrations in Taiwan. The model we design adopts a sequence of recursive networks. This works increased the features of the forecasting system by adopting the datasets extracting both spatial and temporal domains from the collected raw data. This technique of the model is powerful for learning both local and global features better than previous approaches [25, 31, 32, 34]. Firstly, the unstructured data was clustered by considering the corresponding spatial distances among the sensor nodes. Then, the data is clustered into the sectors according to corresponding geographic location and distance for each node. In a sector, every data is converted into a pixel of heat map area by considering the Voronoi area for each node. In addition, the sequence of heat map is processed by using the CNN decoder in order to extract the spatial features for each frame. Further, the Conv-LSTM2D module for the encoder-decoder configuration is implemented in this research, where the visually checking processes are performed to validate the performance and compare the model we propose with the conventional structure. Unlike the conventional approach, the deep learning based on the CRNN model we propose does not require feature manipulation and manual parameter setting because the network parameters are learned directly from the heat maps during training. Therefore, our model does not require re-train when the sensory system is reconfiguring *e.g.* adding, subtracting, and moving sensor nodes. The results of qualitative and quantitative analyses with 12 sequence PM2.5 predictions show that the model we propose works well to make a prediction without involving many atmospheric variables. In the other words, our model minimizes the use of other variables, which is involved in pollutant propagation. Due to capture the spatiotemporal feature only by extracting the sequences of pollutant propagation over time, that the model could minimize the use of other atmospheric variables. The results also show that our model has superior performance in the majority of cases while comparing with traditional 1-dimensional models [21, 22, 25, 34]. In more detail, these better performances are shown in terms of better extraction features, the use of single variables, and better scalability while handling an ever-changing WSN. The risk of using this model is that the number of training parameters is greater than the conventional model. Thus, it would be taken a longer time during training process. However, at the implementation stage, only the inference engine is activated by using fast calculations for real-time system. The training process of our model is only required running one time at the beginning of implementation. Finally, it is understood that developing spatiotemporal features extraction is still a powerful approach with an efficient performance for the model we propose in this paper.

## Acknowledgements

This work was sponsored by the Ministry of Science and Technology (MOST), Taiwan, under Grant No. MOST 107-2221-E-468 -015. This work was also supported in part by Asia University, Taiwan, and China Medical University Hospital, China Medical University, Taiwan, under Grant ASIA-108-CMUH-05 and ASIA-107-CMUH-05. This work was also supported in part by Asia University, Taiwan, UMY, Indonesian, under Grant 107-ASIA-UMY-02.

## References

[1] C.-P. Chio et al., “Health impact assessment of PM2.5 from a planned coal-fired power plant in Taiwan,” Journal of the Formosan Medical Association, vol. 118, no. 11, pp. 1494–1503, Nov. 2019, doi: 10.1016/j.jfma.2019.08.016.

[2] Y. P. Wardoyo, U. P. Juswono and J. A. E. Noor, "Measurements of PM2.5 motor emission concentrations and the lung damages from the exposure mice," 2016 International Seminar on Sensors, Instrumentation, Measurement and Metrology (ISSIMM), Malang, 2016, pp. 99-103, doi: 10.1109/ISSIMM.2016.7803731.

[3] Md. F. Khan, Y. Shirasuna, K. Hirano, and S. Masunaga, “Characterization of PM2.5, PM2.5–10 and PM>10 in ambient air, Yokohama, Japan,” Atmospheric Research, vol. 96, no. 1, pp. 159–172, Apr. 2010, doi: 10.1016/j.atmosres.2009.12.009.

[4] Z. Wei, J. Peng, X. Ma, S. Qiu and S. Wang, "Toward Periodicity Correlation of Roadside PM$_{2.5}$ Concentration and Traffic Volume: A Wavelet Perspective," in IEEE Transactions on Vehicular Technology, vol. 68, no. 11, pp. 10439-10452, Nov. 2019, doi: 10.1109/TVT.2019.2944201.

[5] R. Borge, W. J. Requia, C. Yagüe, I. Jhun, and P. Koutrakis, “Impact of weather changes on air quality and related mortality in Spain over a 25 year period [1993–2017],” Environment International, vol. 133, p. 105272, Dec. 2019, doi: 10.1016/j.envint.2019.105272.

[6] C.-J. Huang, Y. Shen, P.-H. Kuo, and Y.-H. Chen, “Novel spatiotemporal feature extraction parallel deep neural network for forecasting confirmed cases of coronavirus disease 2019,” Socio-Economic Planning Sciences, p. 100976, Nov. 2020, doi: 10.1016/j.seps.2020.100976.

[7] H.-C. Chen, K. T. Putra, S.-S. Tseng, C.-L. Chen, and J. C.-W. Lin, “A spatiotemporal data compression approach with low transmission cost and high data fidelity for an air quality monitoring system,” Future Generation Computer Systems, vol. 108, pp. 488–500, Jul. 2020, doi: 10.1016/j.future.2020.02.032.

[8] X. Li and D. Long, “An improvement in accuracy and spatiotemporal continuity of the MODIS precipitable water vapor product based on a data fusion approach,” Remote Sensing of Environment, vol. 248, p. 111966, Oct. 2020, doi: 10.1016/j.rse.2020.111966.

[9] P. Siarry, A. K. Sangaiah, Y.-B. Lin, S. Mao, and M. R. Ogiela, “Cognitive Big Data Science over Intelligent IoT Networking Systems in Industrial Informatics,” IEEE Trans. Ind. Inf., pp. 1–1, 2020, doi: 10.1109/tii.2020.3024894.

[10] Datasheet from Edimax, AirBox Air Quality Detector with PM2.5, AI-1001W V2, 2017, https://www.edimax.com/edimax/mw/cufiles/files/download/datasheet/AirBox_AI-1001W_V2_Datasheet_English.pdf

[11] J. He, G. Christakos and P. Jankowski, "Comparative Performance of the LUR, ANN, and BME Techniques in the Multiscale Spatiotemporal Mapping of PM2.5 Concentrations in North China," in IEEE Journal of Selected Topics in Applied Earth Observations and Remote Sensing, vol. 12, no. 6, pp. 1734-1747, June 2019, doi: 10.1109/JSTARS.2019.2913380.

[12] R. Yan, J. Liao, J. Yang, W. Sun, M. Nong, and F. Li, “Multi-hour and multi-site air quality index forecasting in Beijing using CNN, LSTM, CNN-LSTM, and spatiotemporal clustering,” Expert Systems with Applications, vol. 169, p. 114513, May 2021, doi: 10.1016/j.eswa.2020.114513.

[13] M. R. Ogiela, F. Palmieri, and M. Takizawa, “Transformative computing approaches for advanced management solutions and cognitive processing,” Information Processing & Management, vol. 57, no. 6, p. 102358, Nov. 2020, doi: 10.1016/j.ipm.2020.102358.

[14] S. Hur, “Short-term wind speed prediction using Extended Kalman filter and machine learning,” Energy Reports, Dec. 2020, doi: 10.1016/j.egyr.2020.12.020.

[15] B. Xu, W. Lin, and S. A. Taqi, “The impact of wind and non-wind factors on PM2.5 levels,” Technological Forecasting and Social Change, vol. 154, p. 119960, May 2020, doi: 10.1016/j.techfore.2020.119960.

[16] M.-Y. Lin et al., “An instantaneous spatiotemporal model for predicting traffic-related ultrafine particle concentration through mobile noise measurements,” Science of The Total Environment, vol. 636, pp. 1139–1148, Sep. 2018, doi: 10.1016/j.scitotenv.2018.04.248.

[17] Y. Zhang, Z. Huang, and S. Wen, “Spatiotemporal variations of in-cabin particle concentrations along public transit routes, a case study in Shenzhen, China,” Building and Environment, vol. 180, p. 107047, Aug. 2020, doi: 10.1016/j.buildenv.2020.107047.

[18] Y. Etzion and D. M. Broday, “Highly resolved spatiotemporal variability of fine particle number concentrations in an urban neighborhood,” Journal of Aerosol Science, vol. 117, pp. 118–126, Mar. 2018, doi: 10.1016/j.jaerosci.2018.01.004.

[19] P. S. S. de Souza et al., “Detecting abnormal sensors via machine learning: An IoT farming WSN-based architecture case study,” Measurement, vol. 164, p. 108042, Nov. 2020, doi: 10.1016/j.measurement.2020.108042.

[20] W. Wang and Y. Guo, "Air Pollution PM2.5 Data Analysis in Los Angeles Long Beach with Seasonal ARIMA Model," 2009 International Conference on Energy and Environment Technology, Guilin, Guangxi, 2009, pp. 7-10, doi: 10.1109/ICEET.2009.468.

[21] M. Oprea, M. Popescu and S. F. Mihalache, "A neural network based model for PM2.5 air pollutant forecasting," 2016 20th International Conference on System Theory, Control and Computing (ICSTCC), Sinaia, 2016, pp. 776-781, doi: 10.1109/ICSTCC.2016.7790762.

[22] J. He, G. Christakos and P. Jankowski, "Comparative Performance of the LUR, ANN, and BME Techniques in the Multiscale Spatiotemporal Mapping of PM2.5 Concentrations in North China," in IEEE Journal of Selected Topics in Applied Earth Observations and Remote Sensing, vol. 12, no. 6, pp. 1734-1747, June 2019, doi: 10.1109/JSTARS.2019.2913380.

[23] P. Du, J. Wang, Y. Hao, T. Niu, and W. Yang, “A novel hybrid model based on multi-objective Harris hawks optimization algorithm for daily PM2.5 and PM10 forecasting,” Applied Soft Computing, vol. 96, p. 106620, Nov. 2020, doi: 10.1016/j.asoc.2020.106620.

[24] B. B. Ustundag and A. Kulaglic, "High-Performance Time Series Prediction With Predictive Error Compensated Wavelet Neural Networks," in IEEE Access, vol. 8, pp. 210532-210541, 2020, doi: 10.1109/ACCESS.2020.3038724.

[25] Y. Tsai, Y. Zeng and Y. Chang, "Air Pollution Forecasting Using RNN with LSTM," 2018 IEEE 16th Intl Conf on Dependable, Autonomic and Secure Computing, 16th Intl Conf on Pervasive Intelligence and Computing, 4th Intl Conf on Big Data Intelligence and Computing and Cyber Science and Technology Congress (DASC/PiCom/DataCom/CyberSciTech), Athens, 2018, pp. 1074-1079, doi: 10.1109/DASC/PiCom/DataCom/CyberSciTec.2018.00178.

[26] D. Wang, Y. Yang and S. Ning, "DeepSTCL: A Deep Spatio-temporal ConvLSTM for Travel Demand Prediction," 2018 International Joint Conference on Neural Networks (IJCNN), Rio de Janeiro, 2018, pp. 1-8, doi: 10.1109/IJCNN.2018.8489530.

[27] S. Song, J. C. K. Lam, Y. Han and V. O. K. Li, "ResNet-LSTM for Real-Time PM2.5 and $PM_{10}$ Estimation Using Sequential Smartphone Images," in IEEE Access, vol. 8, pp. 220069-220082, 2020, doi: 10.1109/ACCESS.2020.3042278.

[28] O. Barron, M. Raison, G. Gaudet, and S. Achiche, “Recurrent Neural Network for electromyographic gesture recognition in transhumeral amputees,” Applied Soft Computing, vol. 96, p. 106616, Nov. 2020, doi: 10.1016/j.asoc.2020.106616.

[29] J. Song, G. Xue, Y. Ma, H. Li, Y. Pan and Z. Hao, "An Indoor Temperature Prediction Framework Based on Hierarchical Attention Gated Recurrent Unit Model for Energy Efficient Buildings," in IEEE Access, vol. 7, pp. 157268-157283, 2019, doi: 10.1109/ACCESS.2019.2950341.

[30] W. Qiao, W. Tian, Y. Tian, Q. Yang, Y. Wang and J. Zhang, "The Forecasting of PM2.5 Using a Hybrid Model Based on Wavelet Transform and an Improved Deep Learning Algorithm," in IEEE Access, vol. 7, pp. 142814-142825, 2019, doi: 10.1109/ACCESS.2019.2944755.

[31] B. Pu, Y. Liu, N. Zhu, K. Li, and K. Li, “ED-ACNN: Novel attention convolutional neural network based on encoder–decoder framework for human traffic prediction,” Applied Soft Computing, vol. 97, p. 106688, Dec. 2020, doi: 10.1016/j.asoc.2020.106688.

[32] T. Li, M. Hua and X. Wu, "A Hybrid CNN-LSTM Model for Forecasting Particulate Matter (PM2.5)," in IEEE Access, vol. 8, pp. 26933-26940, 2020, doi: 10.1109/ACCESS.2020.2971348.

[33] W. Li, L. Zhu, Y. Shi, K. Guo, and E. Cambria, “User reviews: Sentiment analysis using lexicon integrated two-channel CNN–LSTM family models,” Applied Soft Computing, vol. 94, p. 106435, Sep. 2020, doi: 10.1016/j.asoc.2020.106435.

[34] S. Mahajan, H. Liu, T. Tsai and L. Chen, "Improving the Accuracy and Efficiency of PM2.5 Forecast Service Using Cluster-Based Hybrid Neural Network Model," in IEEE Access, vol. 6, pp. 19193-19204, 2018, doi: 10.1109/ACCESS.2018.2820164.

[35] S. Alhirmizy and B. Qader, "Multivariate Time Series Forecasting with LSTM for Madrid, Spain pollution," 2019 International Conference on Computing and Information Science and Technology and Their Applications (ICCISTA), Kirkuk, Iraq, 2019, pp. 1-5, doi: 10.1109/ICCISTA.2019.8830667.

[36] F. Aurenhammer, "Voronoi Diagrams – A Survey of a Fundamental Geometric Data Structure". ACM Computing Surveys. 23 (3): 345–405, 1991. doi:10.1145/116873.116880.